\documentclass{article}

\usepackage[preprint,main]{neurips_2026}
\setcitestyle{numbers,square,comma,compress}
\usepackage[utf8]{inputenc}
\usepackage[T1]{fontenc}

\usepackage{amsmath,amssymb}
\usepackage{amsfonts}
\usepackage{bm}

\usepackage[table]{xcolor} 
\definecolor{citecolor}{HTML}{0071BC}
\usepackage[colorlinks, citecolor=citecolor, linkcolor=red, urlcolor=gray]{hyperref}

\usepackage{url}
\usepackage{booktabs}
\usepackage{multirow}
\usepackage{amsfonts}
\usepackage{wrapfig}
\usepackage{nicefrac}
\usepackage{microtype}
\usepackage{pifont}
\usepackage{makecell}
\usepackage{tabularx}
\usepackage{enumitem}
\usepackage{graphicx}
\usepackage{algorithm}
\usepackage{algpseudocode}
\usepackage{mathtools}
\usepackage{pifont}
\newcommand{\xmark}{\ding{55}}

\usepackage{bm}
\usepackage{caption}

\newlength\savewidth

\title{Trajectory Forcing: Exploiting Diffusion Trajectories \\
for Autoregressive Long Video Generation}

\author{\begin{tabular}[t]{@{}ccc@{}}
    \makebox[0.27\textwidth]{\textbf{Bohan Wang\thanks{Equal contribution.\quad\smash{\textsuperscript{\ensuremath{\dagger}}}Corresponding author.}}} &
    \makebox[0.27\textwidth]{\textbf{Shuo Chen\footnotemark[1]}} &
    \makebox[0.27\textwidth]{\textbf{Yunzhi Li\footnotemark[1]}} \\[4pt]
    \textbf{Zhaozheng Chen} & \textbf{Junzhe Zhang} & \textbf{Hanwang Zhang\textsuperscript{\ensuremath{\dagger}}} \\[6pt]
    \multicolumn{3}{c}{\normalfont Huawei Noah Research Institute} \\[3pt]
    \multicolumn{3}{c}{\normalfont\small\textcolor{black}{\textbf{Project Page:}}~\href{https://wangbohan97.github.io/Trajectory-Forcing/}{\textcolor[HTML]{ED028C}{\nolinkurl{https://wangbohan97.github.io/Trajectory-Forcing/}}}}
  \end{tabular}}

\begin{document}

\maketitle

\begin{abstract}
Autoregressive (AR) modeling is arguably the only way to extend a fixed-length
video diffusion model to longer or even streaming videos. However, existing AR
diffusion methods use only the previously predicted chunk as context for the
next, discarding the intermediate predictions that constitute the chunk's
generation history. In contrast, conventional AR models such as LLMs condition
on the entire history of preceding tokens.
To this end, we propose \textbf{Trajectory Forcing~(TraF)}, an AR diffusion
method that exploits the denoising trajectories of previous video chunks as
context. TraF operates under two training regimes:
(1)~with ground-truth videos, where pseudo-trajectories are constructed from
Gaussian interpolations and re-predicted by the model, and
(2)~with a teacher model, where real trajectories are obtained from the
student's own denoising rollout.
The first regime naturally provides a strong initialization for the second.
On the 5-second VBench benchmark, TraF achieves a Total score of 85.15,
outperforming Self Forcing~(+0.84) and Causal Forcing~(+1.11), with Quality and Semantic
improving simultaneously.
On 30-second generation---6$\times$ the training horizon---TraF
achieves the highest overall quality among all compared methods.
\end{abstract}

\section{Introduction}
\label{sec:intro}

Recent video diffusion models have made rapid progress in generating realistic short videos with complex motion~\cite{sora,wan,veo3}. However, most models still operate within a fixed-length temporal window, limiting both duration and streaming capability. Thus, Autoregressive Diffusion (AR Diffusion), which treats fixed-length diffusion generation as tokens in next-token prediction, is arguably the only approach for jointly modeling temporal sequences of diffusion-generated windows, enabling variable-length generation at inference time for applications such as interactive content creation~\cite{ye2025yan} and world simulation~\cite{genie3}.

\begin{figure}[t]
\centering
\includegraphics[width=\linewidth]{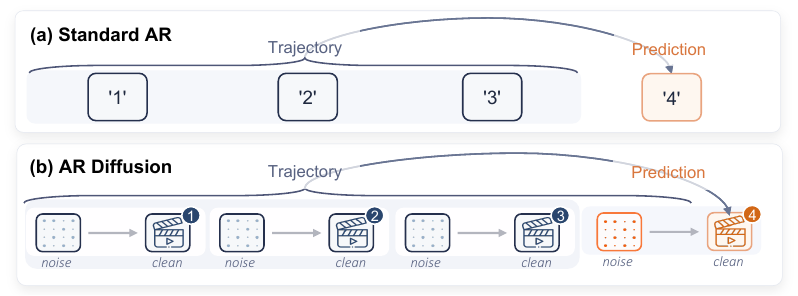}
\caption{\textbf{Standard AR vs.\ AR diffusion.}
(a)~In standard AR~(Eq.~\ref{eq:ar}), each token is atomic and requires no further generation. The next token is naturally conditioned on all preceding tokens.
(b)~In AR diffusion, each chunk is generated through~$T$ denoising steps that form a trajectory. Following Eq.~\eqref{eq:ar_traj}, each denoising step should be conditioned on the full trajectories of all preceding chunks.}
\label{fig:teaser}
\end{figure}

AR models a temporal sequence of length~$N$ via a causal Bayesian
factorization:
\begin{equation}
  p(x_1, \ldots, x_N)
  \;=\; p(x_1)\,\prod_{i=2}^{N}
       p\bigl(x_i \mid x_1, \ldots, x_{i-1}\bigr).
  \label{eq:ar}
\end{equation}
In language models, $x_i$ denotes an atomic token that does not
require further generation~(Figure~\ref{fig:teaser}a).
However, in AR diffusion, the community often assumes that $x_i$
corresponds to a video chunk, overlooking that $x_i$ is no longer
``atomic'' in video generation.
Instead, $x_i$ should consist of $T$ denoising steps,
\textit{i.e.}, $x_i \coloneqq (x_{(i,1)}, \ldots, x_{(i,T)})$, where
$x_{(i,1)}$ is standard Gaussian noise and $x_{(i,T)}$ is the
``clean'' video.
Therefore, a correct formulation of AR diffusion should be:
\begin{align}
  p(x_1, \ldots, x_N)
  &= p\bigl(
       x_{(1,1)}, \ldots, x_{(1,T)},\;\ldots,\;
       x_{(N,1)}, \ldots, x_{(N,T)}
     \bigr)
  \label{eq:ar_traj} \\[2pt]
  &= p(x_{(1,1)})\!\!
     \prod_{(i,t)=(1,2)}^{(N,T)}\;
     p\bigl(x_{(i,t)} \mid
       x_{(1,1)}, \ldots, x_{(i,t-1)}\bigr).
  \nonumber
\end{align}
After this expansion, each element~$x_{(i,t)}$ is atomic, as it
corresponds to a single denoising step and requires no further
internal generation.
Thanks to Eq.~\eqref{eq:ar_traj}, we can immediately know that
existing methods do not consider the \emph{full trajectory} of
each~$x_i$.
For example, in Teacher Forcing~\cite{jin2024pyramidal}, the conditioning on preceding
chunks is reduced to the set of
endpoints~$\{x_{(n,T)}\}_{n<i}$, discarding all the intermediate points~$\{x_{(n,t)}\}_{n<i,\,t<T}$ along the trajectory~(Figure~\ref{fig:teaser}b).
A detailed comparison is presented in Table~\ref{tab:unified}.

After properly formulating AR diffusion in Eq.~\eqref{eq:ar_traj}, we must address error accumulation due to train--test mismatch. A natural approach is to apply Reinforcement Learning (RL, or self-rollout forcing) on top of Supervised Learning (SL, or teacher forcing).
In the SL stage, the model in Eq.~\eqref{eq:ar_traj} is trained using ground-truth videos. During inference, however, ground truth is unavailable, and the model must condition on its own imperfect predictions. This mismatch leads to error accumulation along the autoregressive sequence. In contrast, the RL stage trains the model on its own autoregressive rollouts, with supervision provided only by a trajectory-level reward model, thereby better aligning training and inference conditions. From this perspective, Self Forcing~\cite{huang2025self} can be viewed as a lightweight instance of RL, as it trains on autoregressive rollouts and uses a teacher model to reward the generated videos. We discuss related work for both stages in Section~\ref{sec:related}.

In this paper, we propose \textbf{Trajectory Forcing (TraF)}, which builds on the trajectory-level AR diffusion formulation in Eq.~\eqref{eq:ar_traj} for both the SL and RL stages. Directly conditioning on the full trajectory would multiply the KV-cache size per chunk by the number of denoising steps, significantly increasing memory usage and attention cost, and rendering training impractical on current hardware. Instead, we condition on a single trajectory point per chunk selected at each denoising step (\S\ref{sec:selection}), replacing the static endpoint used by existing methods.
\begin{itemize}[leftmargin=1.5em,itemsep=4pt]

\item \textbf{SL stage}~(\S\ref{sec:pretrain}).
In the SL stage, the model is trained with flow matching~\cite{liu2022flow} on ground-truth videos. Since ground-truth videos do not provide model-generated denoising trajectories, we construct a pseudo-trajectory via linear interpolation. Because this path is a straight line, the entire trajectory can be analytically computed from a \emph{single} velocity prediction (Algorithm~\ref{alg:pretrain}). We show that TraF achieves a total score of 84.39, outperforming Teacher Forcing (\textbf{+0.92}) and Diffusion Forcing (\textbf{+0.32}) (Table~\ref{tab:ablation}).

\item \textbf{RL stage}~(\S\ref{sec:posttrain}).
In the RL stage, the model is trained on its own autoregressive rollouts under reward-model supervision via distribution matching distillation~\cite{yin2024one}. During rollout, the $T$-step denoising process of each chunk naturally produces a trajectory, from which a trajectory point is selected at each denoising step (Algorithm~\ref{alg:posttrain}). TraF achieves a total score of \textbf{85.15}, outperforming Self Forcing by \textbf{0.84} and Causal Forcing by \textbf{1.11} (Table~\ref{tab:main_5s}). For 30-second generation, it achieves the highest total score (\textbf{73.98}) among methods without long-video-specific training (Table~\ref{tab:main_30s}).

\end{itemize}

\section{Related Work}
\label{sec:related}
We organize related AR video diffusion methods by their training stage
and use of the denoising trajectory~(Table~\ref{tab:unified}).

\paragraph{Supervised learning.}
Teacher Forcing~\cite{williams1989learning} is the standard
training stage in current AR video diffusion
systems~\cite{jin2024pyramidal,teng2025magi,yuan2026helios},
where each chunk is conditioned on the ground-truth
endpoints~$\{x_{(n,T)}\}_{n<i}$ of preceding chunks.
Diffusion Forcing~\cite{chen2024diffusion}
and related methods~\cite{song2025history,yin2025slow,wu2025pack,guo2025end,chen2025skyreels}
replace these clean endpoints with corrupted ground-truth versions,
partially mitigating the train--test mismatch.
In both cases, the model still conditions on a single point derived
from ground truth, and the denoising trajectory~$\{x_{(n,\cdot)}\}_{n<i}$ remains unused.

\paragraph{Reinforcement learning.}
Self Forcing~\cite{huang2025self} trains on the model's own
autoregressive rollouts, conditioning each chunk on the
endpoint predictions~$\{x_{(n,T)}\}_{n<i}$ of preceding chunks.
This removes the train--test mismatch but still conditions on a
single trajectory point.
Causal Forcing~\cite{zhu2026causal} extends this stage with
causal ODE initialization and distribution-matching distillation.
Other methods explore rolling diffusion
windows~\cite{liu2025rolling}, minute-scale
generation~\cite{yang2025longlive,cui2025self,zhang2025blockvid},
and diagonal denoising
paths~\cite{liu2026diagdistill}.
The concurrent HiAR~\cite{zou2026hiar} takes a different route by
jointly denoising all chunks in a step-first order, thereby departing
from the sequential AR formulation in Eq.~\eqref{eq:ar_traj}.
Artifacts such as color shift and blurring have been observed
even when the train--test mismatch is
removed~\cite{yuan2026helios,zou2026hiar}, indicating that
the mismatch is not the only source of quality degradation.
As Eq.~\eqref{eq:ar_traj} suggests, each denoising step should
be conditioned on the trajectories of preceding chunks, yet
existing methods still condition on a fixed point regardless of the
denoising step.

\begin{table}[t]
\centering
\small
\renewcommand{\arraystretch}{1.3}
\setlength{\tabcolsep}{6pt}
\begin{tabular*}{\columnwidth}{@{\extracolsep{\fill}} l l l l @{}}
\toprule
  & \textbf{Method}
  & \textbf{SL}
  & \textbf{RL} \\
\midrule
\multirow{3}{*}{\textit{Fixed-point}}
  & Teacher Forcing~\cite{williams1989learning,jin2024pyramidal,teng2025magi,yuan2026helios}
  & GT $x_{(n,T)}$
  & \xmark \\
  & Diffusion Forcing~\cite{chen2024diffusion,song2025history,yin2025slow,wu2025pack,guo2025end,po2025bagger}
  & Corrupted $x_{(n,T)}$
  & \xmark \\
  & Self Forcing~\cite{huang2025self,zhu2026causal,liu2025rolling,yang2025longlive,cui2025self}
  & \xmark
  & Endpoint $x_{(n,T)}$ \\
\midrule
\textit{Trajectory}
  & \textbf{Trajectory Forcing (Ours)}
  & \textbf{Trajectory} $\bm{x}_{(n,\cdot)}$
  & \textbf{Trajectory} $\bm{x}_{(n,\cdot)}$ \\
\bottomrule
\end{tabular*}
\vspace{4pt}
\caption{\textbf{AR diffusion methods categorized by their use
of the denoising trajectory.}
Each column reports the conditioning used under the corresponding
training stage.
\xmark\ indicates that the method does not use that stage.
Prior methods condition on a fixed point regardless of the
denoising step, whereas Trajectory Forcing conditions on the
denoising trajectory~$x_{(n,\cdot)}$.}
\label{tab:unified}
\end{table}

\section{Method}
\label{sec:method}

Trajectory Forcing is motivated by the trajectory-level
formulation in Eq.~\eqref{eq:ar_traj}, which factorizes
autoregressive video diffusion at the level of individual
denoising steps: each~$x_{(i,t)}$ is conditioned on all
preceding trajectory
points~$x_{(1,1)}, \ldots, x_{(i,t-1)}$.
Existing methods reduce this conditioning to the set of
clean endpoints~$\{x_{(n,T)}\}_{n<i}$, discarding all
intermediate points along the trajectory.
Directly conditioning on the full trajectory, however, would
multiply the representations per chunk by the number of
denoising steps, significantly increasing attention
computation and memory usage and making training impractical on
current hardware.
Instead, we condition on a single trajectory point per
chunk selected at each denoising step, replacing the static
endpoint used by existing methods.
We implement this trajectory-level conditioning in two
stages.
Section~\ref{sec:pretrain} constructs pseudo-trajectories
for supervised training on ground-truth videos, and
Section~\ref{sec:posttrain} leverages the real trajectories
that emerge from autoregressive rollouts.

\subsection{SL Stage: Training on Ground Truth}
\label{sec:pretrain}

The model is trained with flow matching~\cite{liu2022flow} on
ground-truth videos, with~$x_0^i$ denoting chunk~$i$.
Eq.~\eqref{eq:ar_traj} requires conditioning on the denoising
trajectory of each preceding chunk, $x_{(n,1)}, \ldots, x_{(n,T)}$,
namely the clean-signal estimates produced at successive
denoising steps.
However, ground-truth videos are not produced by denoising, so
no such trajectory exists.
We address this by constructing a \emph{pseudo-trajectory} via
linear interpolation.
The flow matching forward
process~$x_t = (1{-}t)\,x_0 + t\,\epsilon$ interpolates
linearly between data and noise.
Along this straight line, the oracle velocity~$v^* = \epsilon - x_0$
is constant and recovers the same~$x_0$ at every noise level.
With the learned predictor, each sampled pseudo-state is obtained
from a \emph{single} velocity prediction, and estimates at different
noise levels need not coincide.
These estimates constitute the entire
pseudo-trajectory~(Algorithm~\ref{alg:pretrain}).
The trajectory conditioning prescribed by
Eq.~\eqref{eq:ar_traj} then takes the following form:
\begin{equation}
  p\bigl(x_{(i,t)} \mid
  \underbrace{x_{(1,1)},\, \ldots,\, x_{(1,T)}}_{\text{pseudo-traj.}},\;
  \ldots,\;
  \underbrace{x_{(i{-}1,1)},\, \ldots,\, x_{(i{-}1,T)}}_{\text{pseudo-traj.}}
  \bigr),
  \qquad
  x_{(n,t)} \coloneqq x_t - t \cdot v_\theta,
  \label{eq:sl_form}
\end{equation}
where~$x_{(n,t)}$ is the clean-signal estimate at noise level~$t$,
obtained from a single velocity prediction.

This is distinct from Diffusion
Forcing~\cite{chen2024diffusion}, which conditions directly on a
noise-corrupted version of the ground truth, namely a point on the
\emph{forward} noising path rather than a clean-signal estimate on
the denoising trajectory prescribed by Eq.~\eqref{eq:ar_traj}.
During training, the noise level is sampled uniformly at each
iteration, so the model encounters trajectory points spanning the
full fidelity spectrum, from coarse scene layout at high noise to
fine texture at low noise~\cite{wang2026demystifing}.
Although the pseudo-trajectory is approximate---all points derive
from a single velocity prediction rather than independent denoising
steps---the flow matching loss provides a precise learning signal
by regressing directly to the ground-truth velocity.
The RL stage then replaces this approximation with real
trajectories from the model's own rollout~(\S\ref{sec:posttrain}).

\begingroup
\captionsetup{hypcap=false}
\begin{figure}[t]
\begin{minipage}[t]{0.48\textwidth}
\begin{algorithm}[H]
\caption{SL: Training on Ground Truth}
\label{alg:pretrain}
\small
\begin{algorithmic}[1]
\Require Dataset $\mathcal{D}$, model $v_\theta$
\Loop
  \State Sample $(x_0^{1:N}, \text{prompt})$
    from $\mathcal{D}$
  \State Sample $t \in [0,1]$
  \For{$i = 1, \ldots, N$}
    \Comment{pseudo-traj.\ (no grad)}
    \State $\epsilon_i \!\sim\! \mathcal{N}(0,I)$
    \State Sample $\sigma_i \in [0,1]$
    \State $z_i \!\leftarrow\!
      (1{-}\sigma_i)\,x_0^i + \sigma_i\,\epsilon_i$
    \State $x_{(i)} \!\leftarrow\!
      z_i - \sigma_i \!\cdot\! v_\theta(z_i, \sigma_i, x_{(<i)})$
  \EndFor
  \State Sample $\epsilon'_i \!\sim\! \mathcal{N}(0,I)$
    for each $i$
  \State $x_t^i \!\leftarrow\!
    (1{-}t)\,x_0^i + t\,\epsilon'_i$
  \State $v_i^* \!\leftarrow\! \epsilon'_i - x_0^i$
  \State $\mathcal{L} \!\leftarrow\! \tfrac{1}{N}\!\sum_i
    \bigl\|v_\theta(x_t^i,\, t,\, x_{(<i)})
    - v_i^*\bigr\|^2$
  \State Update $\theta$
\EndLoop
\end{algorithmic}
\end{algorithm}
\end{minipage}
\hfill
\begin{minipage}[t]{0.48\textwidth}
\begin{algorithm}[H]
\caption{RL: Training on Self-Rollouts}
\label{alg:posttrain}
\small
\begin{algorithmic}[1]
\Require Schedule $\{\sigma_1,\ldots,\sigma_T\}$,
  student $G_\phi$, $\gamma$
\Loop
  \State Sample prompt;\;
    $s \!\sim\! \mathrm{Uniform}\{1,\ldots,T\}$
  \For{$i = 1, \ldots, N$}
    \Comment{real traj.\ (rollout)}
    \State $x_{\sigma_1} \sim \mathcal{N}(0, I)$
    \For{$t = 1, \ldots, T$}
      \State $k^* \!\leftarrow\! S_\gamma(p_t)$
        \Comment{Eq.~\ref{eq:selection}}
      \State \textbf{if} $t{=}s$\textbf{:}
        enable gradient
      \State $\hat{x}_0 \!\leftarrow\!
        G_\phi(x_{\sigma_t},\, \sigma_t,\, x_{(<i,\,k^*)})$
      \State $x_{(i,t)} \leftarrow \hat{x}_0$
      \State \textbf{if} $t{=}s$\textbf{:}
        disable gradient
      \State \textbf{if} $t{<}T$\textbf{:}
        ODE step to $\sigma_{t+1}$
    \EndFor
  \EndFor
  \State Compute $\mathcal{L}_{\mathrm{DMD}}$;
    update $\phi$
\EndLoop
\end{algorithmic}
\end{algorithm}
\end{minipage}
\end{figure}
\endgroup

\subsection{RL Stage: Training on Self-Rollouts}
\label{sec:posttrain}

The model is trained on its own autoregressive rollouts via
a student--teacher framework~\cite{huang2025self}.
In this framework, the student~$G_\phi$ generates chunks
autoregressively, and Distribution Matching
Distillation~(DMD)~\cite{yin2024one} provides the training
signal using a frozen teacher score function and a learned
critic~(Algorithm~\ref{alg:posttrain}).

Unlike the SL stage, no trajectory construction is needed.
During rollout, each chunk undergoes $T$~denoising steps,
and each step produces a clean-signal
estimate~$x_{(i,t)}$~($t = 1, \ldots, T$).
These estimates form the real denoising
trajectory~$x_{(i,1)}, \ldots, x_{(i,T)}$, which is precisely the
object required by Eq.~\eqref{eq:ar_traj}.
However, conditioning each step of the current chunk on all
$T$~trajectory points from every preceding chunk requires
attending to $T$~times as many representations.
The resulting increase in attention computation and memory usage
renders training impractical on current hardware.
We therefore select a \emph{single} trajectory point per
preceding chunk at each denoising step, replacing the static
endpoint~$x_{(n,T)}$ used by existing methods. Because different 
steps select different positions along the trajectory, 
selected trajectory points provide context across
steps without the cost of attending to all points at every
step.

\begin{figure}[t]
\centering
\begin{minipage}[c]{0.34\textwidth}
  \centering
  \includegraphics[width=\linewidth]{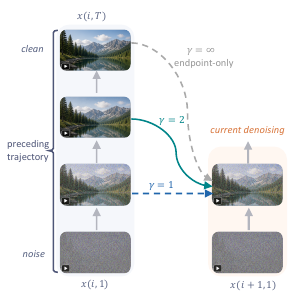}
\end{minipage}
\hfill
\begin{minipage}[c]{0.33\textwidth}
  \centering
  \includegraphics[width=\linewidth]{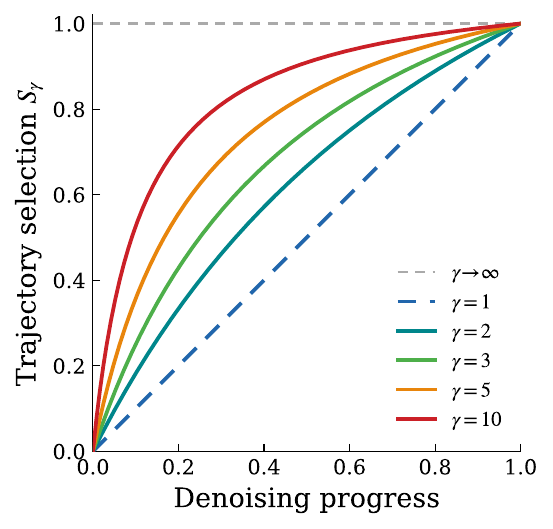}
\end{minipage}
\hfill
\begin{minipage}[c]{0.30\textwidth}
  \centering
  \renewcommand{\arraystretch}{1.3}
  \setlength{\tabcolsep}{3pt}
  \footnotesize
  \begin{tabular}{c c c c}
  \toprule
  $\gamma$ & Quality$\uparrow$ & Semantic$\uparrow$ & Total$\uparrow$ \\
  \midrule
  $\infty$ & 85.04 & 81.07 & 84.24 \\
  $10$     & 85.40 & 81.50 & 84.62 \\
  $5$      & 85.47 & 81.03 & 84.58 \\
  \addlinespace[3pt]
  \rowcolor[gray]{0.93}
  $3$ & \textbf{85.91} & 81.97 & 85.13 \\
 \rowcolor[gray]{0.93}
  $2$      & 85.89 & \textbf{82.20} & \textbf{85.15} \\
  $1$      & 85.49 & 82.10 & 84.82 \\
  \bottomrule
  \end{tabular}
\end{minipage}
\vspace{-4pt}
\caption{\textbf{Trajectory conditioning and $\gamma$ ablation.}
\textit{Left:} Each denoising step of the current chunk
conditions on the preceding chunk's trajectory.
$\gamma$ determines the selected position
($\gamma{\to}\infty$: endpoint-only conditioning, as in existing methods).
\textit{Middle:} Unregularized continuous targets~$S_\gamma$; finite-grid selection is detailed in Appendix~\ref{app:wiener}.
\textit{Right:} The tested finite values of~$\gamma$
outperform endpoint-only conditioning in Total. $\gamma{=}2$--$3$ forms a broad optimum.}
\label{fig:selection}
\end{figure}

\paragraph{Selection function.}
\label{sec:selection}
The selection should reflect the coarse-to-fine structure of
denoising~\cite{wang2026demystifing}. An early step, which
is still resolving scene layout, benefits from a coarser
trajectory point, while a later step, which is refining
textures, benefits from a more refined one.
We formalize this intuition using a Wiener-filter-inspired
SNR criterion~(Appendix~\ref{app:wiener}).
Let $p = 1 - \sigma \in [0,1]$ denote the denoising progress, where
$p = 0$ corresponds to pure noise and $p = 1$ corresponds to the
clean signal.
Requiring the selected trajectory point to lead the current
step by a factor of~$\gamma$ in amplitude SNR yields a
closed-form selection function:
\begin{equation}
  S_\gamma(p) \;=\; \frac{\gamma\, p}
       {1 + (\gamma - 1)\,p},
  \qquad \gamma \geq 1.
  \label{eq:selection}
\end{equation}
At each step with progress~$p$, the model conditions
on~$x_{(\cdot,\,k^*)}$, selected using the discrete rule in Appendix~\ref{app:wiener}.
Eq.~\eqref{eq:selection} defines the continuous target; in the discrete conditioning expressions and Algorithm~\ref{alg:posttrain}, $S_\gamma$ is shorthand for the regularized nearest-point selection specified in Appendix~\ref{app:wiener}.
Motivated by Eq.~\eqref{eq:ar_traj}, we use the following trajectory-point conditioning:
\begin{equation}
  p\bigl(x_{(i,t)} \mid
  \underbrace{x_{(1,k^*)},\; \ldots,\; x_{(i-1,k^*)}}_{\text{real trajectory}}
  \bigr),
  \qquad
  k^* \leftarrow S_\gamma(p_t),
  \label{eq:rl_form}
\end{equation}
where~$k^*$ is the selected trajectory-point index.
This parameterization includes prior methods as special cases
(Figure~\ref{fig:selection}).
Specifically, $\gamma{=}1$ gives step-aligned selection,
while $\gamma \to \infty$ recovers endpoint-only
conditioning~(Table~\ref{tab:unified}).
Empirically, all tested finite~$\gamma$ outperform the endpoint
baseline in Total, with $\gamma \in [2,3]$ forming a broad
optimum~(\S\ref{sec:ablation_gamma}).
All other training components, including gradient truncation to a
single random step and KV detachment across chunks, follow
standard practice~\cite{huang2025self}.

The two stages share the same goal of implementing the
trajectory-level conditioning prescribed by
Eq.~\eqref{eq:ar_traj}.
They differ in trajectory source and supervision. The SL
stage constructs a pseudo-trajectory from ground truth and
trains with flow-matching regression, while the RL stage
operates on real trajectories produced by rollout and trains
with DMD.
This makes the two stages complementary: the SL stage builds
generation capability from precise ground-truth supervision despite
approximate trajectories, while the RL stage aligns this capability
to real inference trajectories via teacher-model supervision.

\section{Experiments}
\label{sec:exp}

\subsection{Experimental Setup}
\label{sec:setup}

\paragraph{Implementation.}
We build on Wan2.1-T2V-1.3B~\cite{wan} with causal attention as the base model, generating $832 \times 480$ videos at 16\,fps.
In the SL stage~(\S\ref{sec:pretrain}), this model is trained with flow matching on ground-truth videos.
In the RL stage~(\S\ref{sec:posttrain}), Wan2.1-T2V-14B serves as the real score network for distribution matching distillation~(DMD)~\cite{yin2024one}.
In the RL stage, we use the same backbone architecture and teacher as Self Forcing~\cite{huang2025self} and Causal Forcing~\cite{zhu2026causal}. TraF is initialized from its SL checkpoint, while the baselines retain their respective original initializations.
Specifically, we use VidProM~\cite{wang2024vidprom} prompts rewritten by Qwen2.5-7B-Instruct~\cite{qwen25}, a batch size of~64, and the Adam optimizer ($\mathrm{lr}{=}2{\times}10^{-6}$, $\beta_1{=}0$, $\beta_2{=}0.999$).
Full training details are provided in Appendix~\ref{app:training}.

\paragraph{Evaluation.}
For short-video evaluation, we adopt the standard VBench~\cite{huang2023vbench} protocol with 946 prompts across all 16 dimensions.
Scores are aggregated into Quality and Semantic following the official VBench weighting, with $\text{Total} = (4 \times \text{Quality} + \text{Semantic}) / 5$ on a 0--100 scale.
For long-video evaluation, we use 128 prompts from MovieGenBench~\cite{polyak2024movie} to generate 30-second videos, with 6 segments of 5\,s each.
Following CausVid~\cite{yin2025slow}, we group the 9 evaluated VBench dimensions into
\emph{Temporal Quality}~(subject consistency, background consistency, temporal flickering, motion smoothness, dynamic degree),
\emph{Frame Quality}~(aesthetic quality, imaging quality), and
\emph{Text Alignment}~(temporal style, overall consistency),
with $\text{Total} = (\text{Temporal Quality} + \text{Frame Quality} + \text{Text Alignment})\,/\,3$.
We also include Quality Drift~\cite{liu2025rolling}, defined as the absolute difference in imaging quality between the first and last 5 seconds of each video, to reflect the severity of error accumulation.
We also report the text-alignment sub-score from VideoScore2~\cite{he2024videoscore2}.
For fair comparison, we equip all block-first methods~(SF, CF, and Ours) with the same sliding-window inference and attention sink~\cite{liu2025rolling,xiao2023efficient}.
The first chunk's KV cache is permanently retained as global context, and subsequent chunks attend to a local window of 21 latent frames.
HiAR~\cite{zou2026hiar}, Rolling Forcing~\cite{liu2025rolling}, and LongLive~\cite{yang2025longlive} use their respective inference architectures.
Evaluation details are provided in Appendix~\ref{app:eval}.
\subsection{Short-Video Results}
\label{sec:main_5s}

\begin{table}[t]
  \centering
  \caption{\textbf{Short-video evaluation on VBench} (946 prompts, all 16 dimensions). Result sources are detailed in Appendix~\ref{app:eval}.
  Throughput and latency are reported in FPS and seconds, respectively. ---: not reported.}
  \label{tab:main_5s}
  \small
  \setlength{\tabcolsep}{4pt}
  \resizebox{\textwidth}{!}{\begin{tabular}{@{} l cc cc ccc @{}}
  \toprule
  \multirow{2}{*}{Method} & \multirow{2}{*}{\#Params} & \multirow{2}{*}{Teacher} & Throughput & Latency & \multirow{2}{*}{Total$\uparrow$} & \multirow{2}{*}{Quality$\uparrow$} & \multirow{2}{*}{Semantic$\uparrow$} \\
  & & & (FPS)$\uparrow$ & (s)$\downarrow$ & & & \\
  \midrule
  \rowcolor[gray]{0.92} \multicolumn{8}{@{}l}{\textit{Bidirectional Video Diffusion}} \\
  LTX-Video~\cite{HaCohen2024LTXVideo}      & 1.9B & \ding{55} & 8.98 & 13.5 & 79.83 & 81.88 & 71.62 \\
  Wan2.1-1.3B~\cite{wan}             & 1.3B & \ding{55} & 0.78 & 103  & 83.37 & 84.30 & 79.65 \\
  \midrule
  \rowcolor[gray]{0.92} \multicolumn{8}{@{}l}{\textit{Autoregressive Video Diffusion}} \\
  MAGI-1~\cite{teng2025magi}                 & 4.5B & \ding{55} & 0.19 & 282  & 78.88 & 81.67 & 67.72 \\
  NOVA~\cite{deng2024nova}                   & 0.6B & \ding{55} & 0.88 & 4.1  & 80.31 & 80.66 & 78.92 \\
  Pyramid Flow~\cite{jin2024pyramidal}       & 2B   & \ding{55} & 6.70 & 2.5  & 80.75 & 83.41 & 70.11 \\
  SkyReels-V2~\cite{chen2025skyreels}        & 1.3B & \ding{55} & 0.49 & 112  & 81.97 & 83.96 & 74.01 \\
  \midrule
  \rowcolor[gray]{0.92} \multicolumn{8}{@{}l}{\textit{Distilled Autoregressive Video Diffusion}} \\
  CausVid~\cite{yin2025slow}                 & 1.3B & Wan-14B & 17.0 & 0.69 & 81.33 & 83.98 & 70.72 \\
  Rolling Forcing~\cite{liu2025rolling}      & 1.3B & Wan-14B & 15.8  & 0.76  & 83.07 & 83.61 & 80.89 \\
  LongLive~\cite{yang2025longlive}           & 1.3B & Wan-14B & 18.0  & 0.65  & 83.09 & 83.69 & 80.69 \\
  HiAR~\cite{zou2026hiar}                   & 1.3B & Wan-14B & ---   & ---   & 83.27 & 84.22 & 79.46 \\
  Causal Forcing~\cite{zhu2026causal}        & 1.3B & Wan-14B & 17.0 & 0.69 & 84.04 & 84.59 & 81.84 \\
  Self Forcing~\cite{huang2025self}          & 1.3B & Wan-14B & 17.0 & 0.69 & 84.31 & 85.07 & 81.28 \\
  \textbf{Trajectory Forcing (Ours)}         & 1.3B & Wan-14B & 15.0 & 0.78 & \textbf{85.15} & \textbf{85.89} & \textbf{82.20} \\
  \bottomrule
  \end{tabular}}
\end{table}

Table~\ref{tab:main_5s} reports VBench scores for short-video generation.
Trajectory Forcing achieves the highest Total~(85.15), Quality~(85.89), and Semantic~(82.20) scores among all compared methods.
\mbox{Compared} with the reported results, it outperforms Self Forcing by +0.84 and Causal Forcing by +1.11 in Total.

The improvement spans both Quality \emph{and} Semantic relative to the reported baseline results.
Self Forcing reports Quality~(85.07) and Semantic~(81.28), while Causal Forcing reports Semantic~(81.84) and Quality~(84.59).
Trajectory Forcing improves both, consistent with our analysis in \S\ref{sec:selection}, where the denoising trajectory provides richer cross-chunk information than a single endpoint and enables gains in both Quality and Semantic without sacrificing either.
Intuitively, the selection function adapts the conditioning fidelity to the current denoising step: early steps receive coarser trajectory points that preserve spatial layout without imposing premature detail, while later steps receive more refined points that guide texture synthesis.
Endpoint-only conditioning provides the same fully detailed reference at every step, which can over-constrain early steps and suppress dynamic content.
The ablation in \S\ref{sec:ablation_corruption} supports our interpretation that this gain stems from trajectory structure rather than generic perturbation.
\subsection{Long-Video Results}
\label{sec:main_30s}

\begin{table}[t]
  \centering
  \caption{\textbf{Long-video evaluation} (128 prompts, 30\,s).
  Temporal Quality, Frame Quality, and Text Alignment are axis aggregates following CausVid~\cite{yin2025slow}. Total is their equal-weighted average.
  The VideoScore2~\cite{he2024videoscore2} column reports the text-alignment sub-score.
  \textcolor[gray]{0.45}{Gray}: long-video-specific training.
  }
  \label{tab:main_30s}
  \small
  \begin{tabular*}{\columnwidth}{@{\extracolsep{\fill}} l cccccc}
  \toprule
  Method & \shortstack{Temporal\\Quality\,$\uparrow$} & \shortstack{Frame\\Quality\,$\uparrow$} & \shortstack{Text\\Align.\,$\uparrow$} & Total$\uparrow$ & \shortstack{Video-\\Score2\,$\uparrow$} & \shortstack{Quality\\Drift\,$\downarrow$} \\
  \midrule
  CausVid~\cite{yin2025slow}            & 83.97 & 62.66 & 58.88 & 68.50 & 3.549 &  2.72 \\
  HiAR~\cite{zou2026hiar}               & 82.82 & 65.32 & 62.42 & 70.19 & 3.758 & 1.60 \\
  Causal Forcing~\cite{zhu2026causal}    & 89.26 & 59.58 & 62.61 & 70.48 & 3.589 & 4.36 \\
  Self Forcing~\cite{huang2025self}      & 85.35 & 61.49 & \textbf{66.33} & 71.06 & 3.719 & 2.21 \\
  \textbf{Trajectory Forcing (Ours)}     & \textbf{90.39} & \textbf{65.46} & 66.08 & \textbf{73.98} & \textbf{3.771} & \textbf{1.38} \\
  \midrule
  \textcolor[gray]{0.45}{Rolling Forcing~\cite{liu2025rolling}}  & \textcolor[gray]{0.45}{82.04} & \textcolor[gray]{0.45}{65.36} & \textcolor[gray]{0.45}{66.77} & \textcolor[gray]{0.45}{71.39} & \textcolor[gray]{0.45}{3.804} & \textcolor[gray]{0.45}{0.56} \\
  \textcolor[gray]{0.45}{LongLive~\cite{yang2025longlive}}       & \textcolor[gray]{0.45}{81.33} & \textcolor[gray]{0.45}{66.07} & \textcolor[gray]{0.45}{67.04} & \textcolor[gray]{0.45}{71.48} & \textcolor[gray]{0.45}{3.801} & \textcolor[gray]{0.45}{0.60} \\
  \bottomrule
  \end{tabular*}
\end{table}

A 30-second video extends the autoregressive chain to six times the training horizon, amplifying the effect of conditioning design on error accumulation.
Among directly comparable methods~(CF, SF, and Ours), Trajectory Forcing achieves the highest Total~(73.98), with the strongest Temporal Quality score~(90.39 vs.\ CF 89.26 and SF 85.35) while remaining competitive on Frame Quality~(65.46).
This is consistent with the motivation in \S\ref{sec:selection}: over a long chain, trajectory context transfers richer information than a single endpoint and better resists the accumulation of visual errors.
Quality Drift further quantifies error accumulation as the absolute imaging-quality gap between the first and last segments.
Trajectory Forcing achieves Quality Drift of 1.38, about 62\% of that of SF~(2.21) and a third of CF~(4.36). The VideoScore2 text-alignment sub-scores are 3.771 for TraF, 3.719 for SF, and 3.589 for CF.
Text Alignment scores are comparable between TraF and SF~(Ours 66.08, SF 66.33), indicating that the primary benefit of trajectory context lies in temporal and visual fidelity rather than text alignment.
This is consistent with the trajectory-level formulation in Eq.~\eqref{eq:ar_traj}: each additional chunk in the autoregressive chain contributes a full trajectory of intermediate predictions, and discarding these in favor of a single endpoint forfeits an increasingly large portion of the generation history.
In the 30-second setting, which extends the chain to six times the training horizon, Trajectory Forcing achieves a higher Total score and lower Quality Drift than SF and CF.

At a broader level, the results suggest that Temporal Quality trades off against Frame Quality and Text Alignment at long horizons.
Causal Forcing achieves high Temporal Quality~(89.26) but the lowest Frame Quality~(59.58), while LongLive and Rolling Forcing show the opposite pattern, achieving higher Frame Quality but substantially lower Temporal Quality, consistent with near-static generation that avoids temporal errors at the cost of dynamic richness.
These two methods benefit from long-video-specific training, which is orthogonal to the conditioning design studied in this work.
Trajectory Forcing achieves the highest Total without such additions and mitigates this trade-off, as trajectory conditioning supplies enough cross-chunk information to keep Temporal Quality high while preserving Frame Quality~(65.46, comparable to RF 65.36 and LongLive 66.07).

\begin{figure}[htp]
  \centering
  \includegraphics[width=\textwidth]{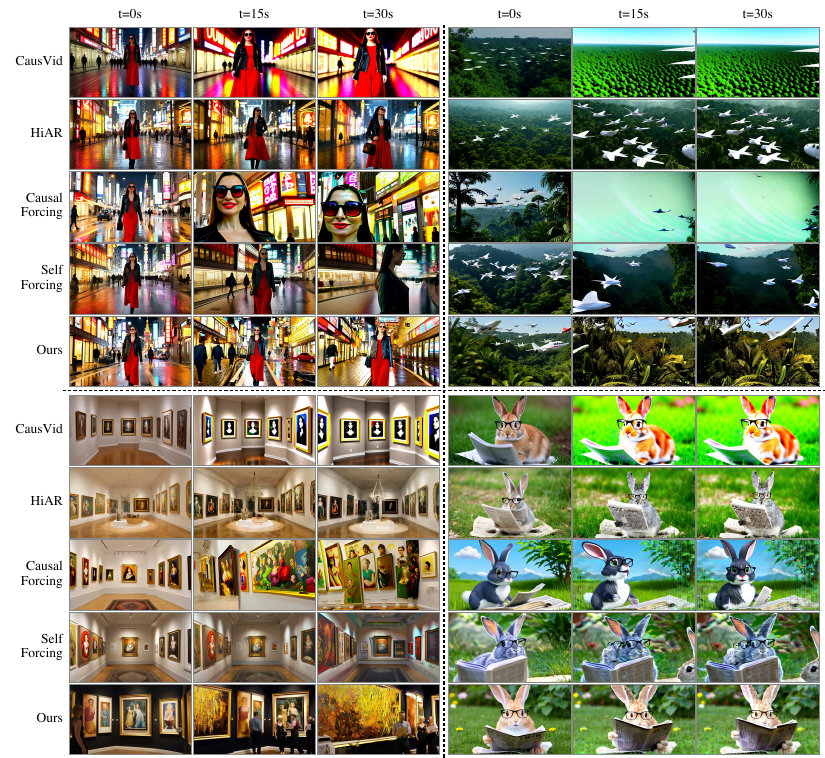}
  \caption{\textbf{Qualitative comparison on 30-second videos.}
  Three frames are sampled at $t{=}0$\,s, $t{=}15$\,s, and $t{=}30$\,s for four representative MovieGen prompts: stylish woman (top-left), paper airplanes (top-right), art gallery tour (bottom-left), and rabbit reading newspaper (bottom-right).
  Trajectory Forcing (Ours) maintains subject coherence and color stability across the full 30~seconds, whereas Causal Forcing exhibits noticeable color drift after $t{=}15$\,s, HiAR produces visually stable but near-static outputs, Self~Forcing shows gradual detail loss in later segments, and CausVid degrades earliest, with pronounced color drift and texture artifacts by $t{=}15$\,s.
}
  \label{fig:qualitative}
\end{figure}
\subsection{Ablation Studies}
\label{sec:ablation}

\begin{table}[t]
\centering
\caption{\textbf{Ablation studies} (5-second VBench).
\textit{Left:} noise vs.\ trajectory comparison, where endpoint predictions with added Gaussian noise of varying magnitude replace trajectory predictions.
\textit{Right:} SL-stage comparison across three conditioning strategies under identical conditions (architecture, data, 20K steps), evaluated with 50-step denoising before distillation.}
\label{tab:ablation}
\small
\begin{minipage}[t]{0.47\textwidth}
  \centering
  \setlength{\tabcolsep}{3.5pt}
  \begin{tabular}{@{} l ccc @{}}
  \toprule
  Conditioning & Total$\uparrow$ & Quality$\uparrow$ & Semantic$\uparrow$ \\
  \midrule
  Gaussian ($\sigma{=}0.22$) & 83.87 & 84.76 & 80.29 \\
  Gaussian ($\sigma{=}0.11$) & 83.95 & 84.92 & 80.08 \\
  \textbf{Trajectory} & \textbf{85.15} & \textbf{85.89} & \textbf{82.20} \\
  \bottomrule
  \end{tabular}
\end{minipage}
\hfill
\begin{minipage}[t]{0.50\textwidth}
  \centering
  \setlength{\tabcolsep}{3.5pt}
  \begin{tabular}{@{} l ccc @{}}
  \toprule
  SL stage & Total$\uparrow$ & Quality$\uparrow$ & Semantic$\uparrow$ \\
  \midrule
  Teacher Forcing & 83.47 & 84.09 & 81.02 \\
  Diffusion Forcing & 84.07 & 84.66 & \textbf{81.68} \\
  \textbf{Trajectory} & \textbf{84.39} & \textbf{85.25} & 80.96 \\
  \bottomrule
  \end{tabular}
\end{minipage}
\end{table}

We conduct three ablations to analyze the selection function, the source of improvement, and the contribution of each training stage.

\paragraph{Selection function.}
\label{sec:ablation_gamma}
Figure~\ref{fig:selection} shows an ablation of the selection parameter~$\gamma$ across the full pipeline.
For this ablation, we train and evaluate a separate model for each tested~$\gamma$.
Every tested finite~$\gamma$ outperforms the endpoint baseline in Total~($\gamma{=}\infty$, Total 84.24), with $\gamma{=}2$--$3$ forming a broad optimum~(Total 85.15 and 85.13).
Step-aligned selection~($\gamma{=}1$, Total 84.82) underperforms this optimum, suggesting that a trajectory point at the same fidelity as the current step provides limited additional information.
The optimum at moderate~$\gamma$ is consistent with the coarse-to-fine principle in \S\ref{sec:selection}, where the selected point should lead the current step in signal quality by a moderate margin.

\paragraph{Noise vs.\ trajectory.}
\label{sec:ablation_corruption}
An alternative explanation is that any degradation of the conditioning signal may suffice to prevent dynamic collapse.
To test this, we replace trajectory predictions with endpoint predictions corrupted by isotropic Gaussian noise at two magnitudes: $\sigma{=}0.22$~(calibrated to match the empirical deviation between trajectory predictions and ground truth) and $\sigma{=}0.11$~(a moderate level).
Table~\ref{tab:ablation}~(left) shows that both Gaussian variants score below the endpoint baseline reported in Figure~\ref{fig:selection}~(Total~84.24), while Trajectory reaches 85.15.
Gaussian corruption is not merely weaker than Trajectory---it is worse than using no perturbation at all.
These Gaussian controls do not reproduce the gain, supporting our interpretation that trajectory predictions provide useful conditioning information beyond the endpoint.

\paragraph{SL stage.}
\label{sec:ablation_pretrain}
The first stage of our pipeline~(\S\ref{sec:pretrain}) trains with pseudo-trajectory conditioning before distillation.
Table~\ref{tab:ablation}~(right) compares three conditioning strategies under identical conditions, evaluated before distillation.
Trajectory achieves the highest Total~(84.39) and Quality~(85.25), outperforming Teacher Forcing by +0.92 and Diffusion Forcing by +0.32 in Total.
Trajectory produces lower Semantic~(80.96) than Diffusion Forcing~(81.68), likely because frequent conditioning on degraded predictions reduces the model's reliance on fine-grained spatial information.
This effect is specific to the pre-distillation stage: in the full pipeline~(Table~\ref{tab:main_5s}), distillation recovers semantic precision, and the final model achieves the highest Semantic~(82.20) among all compared methods.
\subsection{Qualitative Comparison}
\label{sec:qualitative}

Figure~\ref{fig:qualitative} shows keyframes from 30-second videos on representative MovieGen prompts.
Trajectory Forcing maintains subject coherence and color stability across the full 30~seconds, while other methods exhibit varying degrees of color drift, detail loss, or temporal stagnation.
Motion richness and temporal coherence are best assessed via video.

\section{Discussion and Conclusion}
\label{sec:discussion}

Research on autoregressive video diffusion has focused on distillation objectives~\cite{yin2024one,yin2024improved}, training paradigms for exposure bias~\cite{huang2025self,zhu2026causal,chen2024diffusion}, and inference architectures~\cite{liu2026diagdistill,zou2026hiar}.
This work starts from a different observation: in autoregressive diffusion, each chunk is generated through a multi-step denoising process that produces a trajectory of increasingly refined predictions.
The trajectory constitutes the full generation signal of each chunk, yet existing methods reduce it to a single endpoint.
Existing conditioning strategies---teacher forcing, diffusion forcing, and self-forcing---can each be understood as conditioning on ground-truth, forward-corrupted ground-truth, and predicted endpoints, respectively~(Table~\ref{tab:unified}).
We show that conditioning on this trajectory---rather than discarding it---improves both short- and long-video quality while remaining compatible with existing distillation and inference frameworks.

\paragraph{Limitations.}
The current work studies a single base model~(Wan2.1-1.3B) at a single resolution and frame rate.
Verifying generalization across architectures, resolutions, and a broader range of denoising step counts remains an important direction for future work.
Combining trajectory conditioning with long-video training strategies~(\S\ref{sec:main_30s}) is a natural extension that we have not yet explored.
Under the reported setting, TraF improves quality at lower throughput than Self Forcing (15.0 vs.\ 17.0 FPS); we have not evaluated an exactly matched-throughput operating point.

\paragraph{Conclusion.}
Trajectory Forcing implements trajectory-level conditioning in both training stages: pseudo-trajectories constructed from ground-truth videos in the SL stage, and real trajectories produced by autoregressive rollout in the RL stage.
The method achieves the highest scores on both short-video~(85.15 Total) and long-video~(73.98 Total) benchmarks among directly comparable methods.
On 30-second generation, which extends six times beyond the training horizon, Trajectory Forcing achieves the lowest Quality Drift among directly comparable methods~(Table~\ref{tab:main_30s}).
These results suggest that the denoising trajectory is a rich and previously untapped signal for autoregressive video generation.

\begin{ack}
This work received no external funding.
The authors declare no competing financial interests beyond their listed primary affiliations.
\end{ack}

\newpage
\appendix

\section{Wiener-Inspired Selection Rule}
\label{app:wiener}
We derive the selection function $S_\gamma$ using an SNR-guided heuristic motivated by the Wiener filtering perspective outlined in \S\ref{sec:selection}.

\paragraph{Setup.}
Under the flow matching forward process $x_t = (1-t)\,x_0 + t\,\epsilon$ with $\epsilon \sim \mathcal{N}(0, I)$, the minimum mean-square error (MMSE) estimate of $x_0$ given $x_t$ is $\hat{x}_0 = \mathbb{E}[x_0 \mid x_t]$.
Under a linear-Gaussian signal model, this estimate acts as a Wiener filter~\cite{wiener1949extrapolation} whose effective bandwidth is determined by the signal-to-noise ratio.

Following the standard convention, the signal-to-noise ratio at noise level $t$ is the ratio of signal variance to noise variance:
\begin{equation}
  \mathrm{SNR}(t) \;\triangleq\; \frac{\alpha_t^2}{\sigma_t^2} \;=\; \frac{(1-t)^2}{t^2}.
\end{equation}

\paragraph{From variance to amplitude.}
While $\mathrm{SNR}$ is canonically defined in power (variance) units, we parameterize the lookahead in amplitude (standard deviation) units.
We use the v-prediction parameterization, $v_t = \dot{\alpha}_t\,x_0 + \dot{\sigma}_t\,\epsilon$.
The ratio $\alpha_t / \sigma_t$ compares the signal and noise scaling coefficients in the forward interpolation.
We adopt this amplitude SNR as a heuristic measure of how much one denoising step ``leads'' another:
\begin{equation}
  \sqrt{\mathrm{SNR}(t)} \;=\; \frac{\alpha_t}{\sigma_t} \;=\; \frac{1-t}{t}.
\end{equation}
In terms of denoising progress $p = 1-\sigma \in [0,1]$, this becomes $\sqrt{\mathrm{SNR}(p)} = p/(1-p)$.

\paragraph{Derivation.}
At denoising step $j$ with progress $p_j$, the current generation operates at amplitude SNR $p_j/(1-p_j)$.
We seek a context trajectory point whose amplitude SNR leads the current step by a factor $\gamma \geq 1$:
\begin{equation}
  \sqrt{\mathrm{SNR}(p_c)} \;=\; \gamma \cdot \sqrt{\mathrm{SNR}(p_j)}.
\end{equation}
Substituting and cross-multiplying:
\begin{equation}
  \frac{p_c}{1-p_c} \;=\; \gamma \cdot \frac{p_j}{1-p_j}
  \qquad\Longrightarrow\qquad
  p_c \;=\; \frac{\gamma\, p_j}{1 + (\gamma - 1)\,p_j}.
\end{equation}
This yields the selection function $S_\gamma(p) = \gamma p\,/\,[1 + (\gamma-1)p]$ stated in Eq.~\ref{eq:selection}.

\paragraph{Properties.}
\textbf{No lookahead ($\gamma = 1$).}
$S_1(p) = p$, so the context step $p_c$ is exactly equal to the current step $p_j$.
There is no ``lead'' or ``lookahead'' between context and query, and the scheme reduces to step-aligned conditioning.

\textbf{Degenerate lookahead ($\gamma \to \infty$).}
$S_\gamma(p) \to 1$ for all $p > 0$, so the context step $p_c$ is pushed all the way to $1$, corresponding to the fully clean signal $x_0$, as in Self Forcing~\cite{huang2025self} and Causal Forcing~\cite{zhu2026causal}.

\textbf{Balanced lookahead ($\gamma = 2$).}
This setting provides a moderate lead: for $0 < p < 1$, the continuous target progress $S_2(p)$ has nominal amplitude SNR exactly double that of the current progress $p$, without forcing endpoint-only conditioning.
The iterative refinement structure is preserved.
Consistent with this analysis, $\gamma \in [2,3]$ forms a broad empirical optimum (Figure~\ref{fig:selection}).

\paragraph{Discrete trajectory selection.}
For finite-grid selection, the regularized target amplitude SNR and progress are
\[
  r_c = \gamma\frac{p}{1-p} + (\gamma-1)\delta_T,
  \qquad p_c = \frac{r_c}{1+r_c}.
\]
We select the available trajectory point whose progress is nearest to~$p_c$.
For uniform pre-shift noise levels $1-j/T$, $j=0,\ldots,T-1$, with shift factor~$5$, we set $\delta_T=2/[3(10T-9)]$.
For a fixed denoising grid and~$\gamma$, the selected indices can be precomputed, while the retrieved trajectory predictions remain sample-dependent.

\section{Training Details}
\label{app:training}

Our training pipeline consists of two stages, both built on the open-source Wan2.1-T2V-1.3B~\cite{wan} codebase with causal attention.
For the SL stage, we use a filtered subset of OpenVid-1M~\cite{nan2024openvid}. The RL settings and model initializations are described in \S\ref{sec:setup}.
Table~\ref{tab:app_config} summarizes the configuration.

\begin{table}[h]
  \centering
  \caption{Training configuration. The settings above the first divider refer to the RL stage.}
  \label{tab:app_config}
  \small
  \begin{tabular}{@{} ll @{}}
  \toprule
  \textbf{Component} & \textbf{Setting} \\
  \midrule
  Base model & Wan2.1-T2V-1.3B (causal attention) \\
  Video resolution & $832 \times 480$, 16\,fps, 81 frames (5\,s) \\
  RL prompts & VidProM~\cite{wang2024vidprom} (${\sim}$250K after filtering) \\
  RL prompt rewriting & Qwen2.5-7B-Instruct~\cite{qwen25} \\
  Optimizer & Adam ($\mathrm{lr}{=}2{\times}10^{-6}$, $\beta_1{=}0$, $\beta_2{=}0.999$) \\
  Batch size & 64 \\
  Noise schedule & Flow matching with time-step shifting ($k{=}5$) \\
  Inference steps & 4 (following Self Forcing~\cite{huang2025self}) \\
  \midrule
  \multicolumn{2}{@{}l}{\textit{Stage 1: Ground-truth supervision (\S\ref{sec:pretrain})}} \\
  Training steps & 20K \\
  Supervision & Flow matching velocity loss \\
  Conditioning & Pseudo-trajectory (\S\ref{sec:pretrain}) \\
  \midrule
  \multicolumn{2}{@{}l}{\textit{Stage 2: Teacher supervision (\S\ref{sec:posttrain})}} \\
  Teacher (real score) & Wan2.1-T2V-14B \\
  Critic (fake score) & Initialized from 1.3B base model \\
  Training steps & 2K \\
  Supervision & DMD~\cite{yin2024one} (reverse KL) \\
  Conditioning & Trajectory (\S\ref{sec:posttrain}) \\
  Gradient truncation & Single random step per rollout \\
  \bottomrule
  \end{tabular}
\end{table}

\paragraph{Noise schedule.}
Following Wan2.1, we adopt the flow matching framework with time-step shifting $t'(k, t) = kt\,/\,(1 + (k-1)t)$ and shift factor $k = 5$.
The forward process is $x_t = (1-t)\,x_0 + t\,\epsilon$ with $\epsilon \sim \mathcal{N}(0, I)$ and $t \in [0, 1]$.
The model predicts velocity $v_\theta(x_t, t)$, and the clean-signal estimate is $\hat{x}_0 = x_t - t \cdot v_\theta(x_t, t)$.

\paragraph{Prompt preprocessing.}
In the RL stage, we use the VidProS subset of VidProM~\cite{wang2024vidprom}, filtering prompts shorter than 20 characters, containing command-line arguments, or with NSFW probability above 0.01.
This yields approximately 250K prompts, which are rewritten with Qwen2.5-7B-Instruct~\cite{qwen25} using the system prompt from the Wan2.1 release.
VBench evaluation prompts are similarly rewritten for consistency.

\paragraph{Auxiliary predictor diagnostic.}
On 32 clips (192 chunk-samples, chunks 2--7), we fix the prompt, preceding ground-truth context, and noise~$\epsilon$ within each chunk, and separately query the Stage-1 predictor at $\sigma\in\{0.25,0.50,0.625,0.833,0.9375\}$.
The mean pairwise RMS distances between these estimates, normalized by the RMS of~$x_0$, range from 3.7\% to 19.7\%.
This diagnostic is separate from the fixed-velocity analytical construction in Algorithm~\ref{alg:pretrain}.

\paragraph{DMD details.}
In Stage~2, we write the DMD gradient schematically, omitting the weighting and noise-schedule factors:
$\nabla_\phi \mathcal{L}_{\mathrm{DMD}} = -\mathbb{E}_{t,\hat{x}_t}[(s_{\mathrm{real}}(\hat{x}_t, t) - s_{\mathrm{fake}}(\hat{x}_t, t))\,\partial G_\phi / \partial \phi]$,
where $s_{\mathrm{real}}$ is the frozen Wan2.1-14B teacher and $s_{\mathrm{fake}}$ is a learned critic initialized from the 1.3B base model.
We use the DMD objective adopted by Self Forcing~\cite{huang2025self}.
The inference window covers the full five-second training context.
Under the default four-step Stage-2 recipe, trajectory-cache maintenance adds an estimated 6\% to training time relative to the matched endpoint-only loop, with no additional denoiser evaluations.

\clearpage
\section{Evaluation Details}
\label{app:eval}

\paragraph{5-second VBench.}
We evaluate all 16 VBench~\cite{huang2023vbench} dimensions using 946 prompts.
Quality score aggregates 7 dimensions (subject consistency, background consistency, temporal flickering, motion smoothness, dynamic degree, aesthetic quality, imaging quality), where dynamic degree receives weight 0.5 and all others receive weight 1.0.
Semantic score aggregates 9 dimensions (object class, multiple objects, human action, color, spatial relationship, scene, temporal style, overall consistency, appearance style), all equally weighted.
Total is computed as $(4 \times \text{Quality} + \text{Semantic})\,/\,5$ on a 0--100 scale.
Models evaluated by us use rewritten prompts when supported.

Results for LTX-Video, Wan2.1-1.3B, MAGI-1, NOVA, Pyramid Flow, SkyReels-V2, CausVid, and Causal Forcing are taken from Causal Forcing~\cite{zhu2026causal}. Self Forcing results are taken from its original paper~\cite{huang2025self}. HiAR, Rolling Forcing, and LongLive results are obtained from our evaluation.

\paragraph{Repeatability.}
In each run, the endpoint-only baseline ($\gamma=\infty$) and TraF start from the same SL checkpoint.
Their VBench Total scores are 84.24/85.15 in the first run and 84.33/85.13 in the second run, respectively.

\paragraph{Ten-step transfer.}
We retrain matched trajectory-conditioned and endpoint-only DMD models at $T=10$ under two training seeds.
Within each pair, both arms share the initialization, optimization settings, uniform ten-step schedule, and frozen Wan2.1-1.3B real-score teacher; both use a 1,000-iteration budget and the final checkpoint.
We retain $\gamma=2$ without retuning.
On the full 946-prompt VBench protocol, endpoint-only DMD/TraF Total scores are 83.26/84.33 and 83.44/84.45, giving a mean gain of +1.04 (paired prompt-bootstrap 95\% CI: [+0.17, +1.90]).
These experiments use a 1.3B teacher rather than the 14B teacher in the four-step setting, so we compare matched within-setting gains rather than absolute scores across step counts.

\paragraph{Non-distilled ten-step inference.}
In a separate evaluation, we hold the non-distilled SL checkpoint fixed and vary only~$\gamma$ under a standard ten-step Flow Matching ODE solver.
On the full 946-prompt VBench protocol, Total scores are 83.51, 83.90, 83.77, and 83.20 for $\gamma=1,2,3,\infty$, respectively.
The default~$\gamma=2$ improves over endpoint conditioning by +0.70 (paired prompt-bootstrap 95\% CI: [+0.17, +1.17]).

\paragraph{30-second evaluation.}
We generate 30-second videos using the first 128 prompts, in file order, from the public 1,003-prompt MovieGenBench~\cite{polyak2024movie} list distributed with Self Forcing~\cite{huang2025self}.
Each 30-second video consists of 126 latent frames, split into 6 segments of 5 seconds each.
We evaluate 9 VBench dimensions in \texttt{custom\_input} mode on each segment and average across segments.
The remaining 7 semantic dimensions (object class, multiple objects, human action, color, spatial relationship, scene, appearance style) are excluded because they require dimension-specific prompt templates not available in the MovieGen prompt set.
Following CausVid~\cite{yin2025slow}, the 9 evaluated dimensions are grouped as
\textbf{Temporal Quality}~(5 dims: subject consistency, background consistency, temporal flickering, motion smoothness, dynamic degree),
\textbf{Frame Quality}~(2 dims: aesthetic quality, imaging quality), and
\textbf{Text Alignment}~(2 dims: temporal style, overall consistency),
with $\text{Total} = (\text{Temporal Quality} + \text{Frame Quality} + \text{Text Alignment})\,/\,3$.
All raw scores are normalized to a 0--100 scale using the same normalization constants as Causal Forcing~\cite{zhu2026causal}.
Treating prompts as the independent units, we perform 10,000 paired bootstrap resamples over the 128 prompts after aggregating each prompt's six segments.
The 30-second Total gain over Self Forcing is +2.92 (95\% CI: [+2.06, +3.78]), quantifying uncertainty from prompt sampling.

\begin{table}[ht]
  \centering
  \caption{Score differences and 95\% paired prompt-bootstrap confidence intervals for 30-second evaluation.}
  \label{tab:app_30s_bootstrap}
  \small
  \begin{tabular}{@{}lcc@{}}
  \toprule
  Metric & TraF $-$ Self Forcing & TraF $-$ Causal Forcing \\
  \midrule
  Total & $+2.92\,[+2.06,+3.78]$ & $+3.50\,[+2.76,+4.26]$ \\
  Temporal Quality & $+5.04\,[+3.62,+6.47]$ & $+1.13\,[+0.23,+2.01]$ \\
  Frame Quality & $+3.97\,[+3.07,+4.89]$ & $+5.88\,[+5.01,+6.75]$ \\
  Text Alignment & $-0.25\,[-2.03,+1.59]$ & $+3.47\,[+1.77,+5.26]$ \\
  \bottomrule
  \end{tabular}
\end{table}

\paragraph{Inference configuration for 30-second videos.}
All block-first methods (SF, CF, and Ours) use identical sliding-window inference with \texttt{local\_attn\_size=21} and \texttt{sink\_size=3}, where the first chunk is permanently retained in the KV cache as global context~\cite{xiao2023efficient,liu2025rolling}.
EMA weights are used for all methods.
HiAR uses its native step-first denoising with pipeline parallelism.
Rolling Forcing uses its rolling diffusion window.
LongLive uses its short-window attention with frame sink.

\paragraph{VideoScore2.}
We use VideoScore2~\cite{he2024videoscore2} with the Qwen2.5-VL backbone to evaluate text alignment.
Each 5-second segment is scored independently, and the text-alignment sub-score is averaged across all 6 segments.

\section{Broader Impacts}
\label{app:impact}

Trajectory Forcing may support creative production, animation, and educational visualization through the synthesis of extended video sequences.
These capabilities also create risks: synthetic videos could be used to fabricate events, misrepresent real people, or spread misleading visual narratives.
Longer sequences may make fabricated content more persuasive when detached from its original context.
Responsible use calls for clear identification of synthetic content and careful consideration of consent, privacy, and intellectual property.
We encourage downstream users to assess these risks in their intended applications and avoid presenting generated examples as documentary evidence.

\end{document}